\documentclass{article} 
\usepackage[T1]{fontenc}
\usepackage[utf8]{inputenc}
\usepackage{iclr2015,times}
\usepackage{hyperref}
\usepackage{url}
\usepackage[american]{babel}
\usepackage[autostyle=true]{csquotes}
\usepackage[shortcuts]{extdash}
\usepackage{microtype}
\usepackage{comment}
\usepackage{caption}

\usepackage{standalone} 
\usepackage{tikz}
\usetikzlibrary{calc,patterns,backgrounds,arrows,positioning,spy,shapes,shadows}
\usepackage{pgfplots,pgfplotstable}

\usepackage{bm,times}
\usepackage{verbatim}

\usepackage{graphicx}
\usepackage{caption}
\usepackage{subcaption}
\usepackage{tikz}
\usepackage{amsthm}
\usepackage{amssymb}
\usepackage{amsmath}
\usepackage{xcolor}
\usepackage{colortbl}

\newcommand{\mytitle}{Grounding Hierarchical Reinforcement Learning Models for Knowledge Transfer}

\newcommand{\myname}{Mark Wernsdorfer}
\newcommand{\authors}{\myname, Ute Schmid}
\newcommand{\keywords}{Transfer Learning, Deep Reinforcement Learning, Representation Learning, Model-based Reinforcement Learning}

\usepackage{bookmark}
\hypersetup{%
    pdftitle={\mytitle},%
    pdfauthor={\authors},%
    pdfsubject={\keywords},%
    pdfkeywords={\keywords},%
    draft=false,%
    colorlinks=false,%
    pdfborder={0 0 0},%
    pdfstartview=,%
    pdfdisplaydoctitle=true,%
}

\usepackage{cleveref}

\iclrconference

\title{\mytitle}

\author{
\authors \thanks{ \url{http://www.uni-bamberg.de/kogsys/}} \\
Cognitive Systems Group\\
Faculty of Information Systems and Applied Computer Sciences\\
Otto-Friedrich-Universität Bamberg\\
An der Weberei 5, 96047 Bamberg, Germany \\
\texttt{\{mark.wernsdorfer, ute.schmid\}@uni-bamberg.de}
}

\begin{document}
\pgfplotsset{max space between ticks=50pt}
\tikzset{every mark/.append style={scale=.5}}

\maketitle

\begin{abstract}
Methods of \emph{deep machine learning} enable to to reuse low\-/level representations efficiently for generating more abstract high\-/level representations. Originally, deep learning has been applied passively (e.g., for classification purposes). Recently, it has been extended to estimate the value of actions for autonomous agents within the framework of \emph{reinforcement learning} (RL). Explicit models of the environment can be learned to augment such a value function. Although \enquote{flat} connectionist methods have already been used for model\-/based RL, up to now, only model\-/free variants of RL have been equipped with methods from deep learning. We propose a variant of \emph{deep model\-/based RL} that enables an agent to learn arbitrarily abstract hierarchical representations of its environment. In this paper, we present research on how such hierarchical representations can be grounded in sensorimotor interaction between an agent and its environment.
\end{abstract}

\section{Introduction} \label{sec:intro}
Machine learning algorithms are derived from models of the problem\-/to\-/be\-/solved. In RL, problems are modeled in the form of \emph{Markov decision processes} (MDPs). A MDP can be defined as the 4-tuple $(S, A, \mathcal{T}, \mathcal{R}) $. The first element $S $ describes the \emph{set of wold states}---in most cases it is interpreted as the agent's sensory perception. The second element $A $ describes the \emph{set of possible actions.} The \emph{transition function} determines transitions in state space. It has the general form $\mathcal{T}: S \times A \rightarrow S $. A particular action in a particular state causes a particular successor state. Lastly, the \emph{reward function} gives a scalar value which acts as reward. It indicates whether a particular action in a particular state is to be repeated or avoided. It has the general form $\mathcal{R}: S \times A \rightarrow R $ where usually $R = \mathbb{Z} $.

Although $S $ might be chosen in a way that captures the whole state of the world, for realistic applications, it must be restricted to cover only a small subset of the statements that are currently true about the world. Consider the following examples.
\begin{enumerate}
\item Observable world state
\begin{itemize}
\item The agent perceives that it is in position (2, 13).
\item It moves north to enter sensor state (2, 12).
\end{itemize}
\item Hidden world state
\begin{itemize}
\item The agent perceives a wall in front of it.
\item It turns around to enter a sensor state that suggests free passing.
\end{itemize}
\end{enumerate}

To model the second case, the agent's restricted perception must be taken into account. To cover this, MDPs can be generalized into \emph{partially observable MDPs} (POMDPs). They can be defined as the 6-tuple $(S, A, \mathcal{T}, \mathcal{R}, \Omega, \mathcal{O}) $. In addition to the four elements above, these models also comprise the \emph{set of possible observations} $\Omega $---possibly distinct from the set of world states---and the \emph{observation function} which has the general form $\mathcal{O}: S \times A \rightarrow \Omega $. A particular action in a particular world state causes a particular observation. \citep{Kaelbling1998}

POMDPs can describe a plethora of tasks for RL agents. If the set of world states is accessible to the agent, it can infer a particular \emph{belief} about what the current world state is. Indications for the current world state can be inferred from previous beliefs and those world states, that recently observed transitions match with. The result is a belief in the form of a probability distribution over all world states.

As a consequence, the resulting \emph{belief MDP} describes transitions between continuous probability distributions instead of discrete states. To generate probability distributions over wold states, however, these world states need to be known beforehand. Providing the set of world states a priori implies two fatal consequences.

\subsection{Semantic Load of Representations} \label{ssec:semantic}
The first problem concerns the \emph{semantic load} of predefined representations. The set of world states, and therefore the beliefs inferred from them, can be regarded as \emph{representations of the environment.} To provide an agent with such a fixed set of representations, however, means to inject considerable amounts of knowledge into the system.

It tacitly implies that these representations are appropriate for solving the POMDP at hand. Having appropriate representations implies knowledge about how those representations are to be used. To know how to use representations, on the other hand, means to already know how to handle the task at hand.

\citet{Bengio2013} mention the necessary trade\-/off between the \enquote{richness} of representations and the effort necessary to process them. \citet{Diuk2008} show that the selection of representations has immense influence on the performance of RL algorithms. More specifically, they show that providing \emph{semantically rich} representations (e.g. representations of objects) can simplify a given task considerably compared to \emph{semantically sparse} representations (e.g. representations of world states).

\subsection{Transfer of Knowledge}
The second problem is that \emph{knowledge transfer} with objective inputs requires elaborate methods for joining discrete data. Partial observability is frequently avoided by providing the world state as sensory input. In any (stochastically) determinate world, complete world states enable to learn an interaction policy which optimizes the cumulative reward received over time. The detachment of objective \enquote{snapshots} of the whole world, however, masks the actually relevant information in interaction data.

The first example of observable world states is such a case of \emph{objective interaction.} Knowledge transfer with objective data is successfully achieved if the equivalence of seemingly different states is discovered. It might be discovered, for example, that column 2 in a grid world is always devoid of obstacles. Therefore, state transitions to the north in row 2 can be modeled by approximating the transition function $\mathcal{T} $ with $T_2((x, y), a_n) = (x, y-1) $. All states in column 2 can be pooled according to their successor state when the agent moves to the north. The data structure implementing $T_2 $ can be regarded as \emph{a high\-/level representation} of column 2.

The second example of hidden world states describes a case of \emph{subjective interaction.} The agent has no access to absolute information about its relation to the environment. Assuming the existence of an objective agent position, its perception must therefore be regarded as ambiguous. Every single location in column 2 might be perceived as identical, although, from an objective perspective, they differ quite obviously in their value of $x $, the absolute position of the agent on a vertical axis.\footnote{This concerns not only perception. In relative interaction, actions also depend on hidden world states: moving north, east, south, or west is not part of the agent's motor capabilities. Instead, its absolute direction of movement depends on its current orientation, which is not directly observable. The objective direction of movement is not only for the agent to decide.}

Depending on the interaction paradigm, the problem of knowledge transfer takes two contrastive forms. Objective interaction makes it necessary to \emph{analyze} states which appear to the agent as if they were different and \emph{pool} them to recognize their \emph{similarity.} In return, the agent does not have ambiguous perceptions from an objective observer's perspective.

Subjective interaction, on the other hand, makes it necessary to \emph{synthesize} states which appear to the agent as if they were identical and \emph{differentiate} them to recognize their \emph{distinctness.} In return, knowledge transfer comes for free: although the agent might be in objectively different situations, as long as there is no reason for differentiation, knowledge which has been been acquired from identical perceptions and actions is applied effortlessly, in fact even unnoticed by the agent.

\begin{figure}
\centering
\begin{subfigure}[b]{.48\textwidth}
	\centering
\begin{tikzpicture}[scale=.5, even odd rule]\footnotesize
    \pgfmathsetmacro{\xone}{-.5}
    \pgfmathsetmacro{\xtwo}{7.5}
    \pgfmathsetmacro{\yone}{-.5}
    \pgfmathsetmacro{\ytwo}{4.5}
    \pgfmathsetmacro{\xs}{1}
    \pgfmathsetmacro{\ys}{1}

    \begin{scope}<+->;
        \draw[step=1cm,gray,very thin, opacity=.5] (\xone,\yone) grid (\xtwo,\ytwo);
    \end{scope}

    \begin{scope}
        \draw[line width=.05cm,color=red!30,cap=round,join=round,->] (1.5,1.9)--(1.5,2.5)--(3.5,2.5)--(3.5,1.5)--(5.1,1.5);

        \draw[black, pattern=north east lines] (0, 0) rectangle (7, 4) (1,3)--(1,1)--(2,1)--(2,2)--(3,2)--(3,1)--(6,1)--(6,2)--(4,2)--(4,3)--(1,3);

        \draw[black] (1.5, 1.5) node {$S$};
        \draw[black] (5.5, 1.5) node {$G$};
    \end{scope}

    \begin{pgfonlayer}{background}
	    \fill [white] (\xone, \yone) rectangle (\xtwo, \ytwo);
    \end{pgfonlayer}
\end{tikzpicture}
	\caption{Training environment: source of knowledge.} \label{fig:training_env}
\end{subfigure}
~
\begin{subfigure}[b]{.48\textwidth}
	\centering
\begin{tikzpicture}[scale=.5, even odd rule]\footnotesize
    \pgfmathsetmacro{\xone}{-.5}
    \pgfmathsetmacro{\xtwo}{11.5}
    \pgfmathsetmacro{\yone}{-.5}
    \pgfmathsetmacro{\ytwo}{4.5}
    \pgfmathsetmacro{\xs}{1}
    \pgfmathsetmacro{\ys}{1}

    \begin{scope}<+->;
        \draw[step=1cm,gray,very thin, opacity=.5] (\xone,\yone) grid (\xtwo,\ytwo);
    \end{scope}

    \begin{scope}
        \draw[line width=.05cm,color=red!30,cap=round,join=round,->] (1.5,1.9)--(1.5,2.5)--(3.5,2.5)--(3.5,1.5)--(5.5,1.5)--(5.5,2.5)--(7.5,2.5)--(7.5,1.5)--(9.1,1.5);

        \draw[black, pattern=north east lines] (0, 0) rectangle (11, 4) (1,3)--(1,1)--(2,1)--(2,2)--(3,2)--(3,1)--(6,1)--(6,2)--(7,2)--(7,1)--(10,1)--(10,2)--(8,2)--(8,3)--(5,3)--(5,2)--(4,2)--(4,3)--(1,3);

        \draw[black] (1.5, 1.5) node {$S$};
        \draw[black] (9.5, 1.5) node {$G$};
    \end{scope}

    \begin{pgfonlayer}{background}
	    \fill [white] (\xone, \yone) rectangle (\xtwo, \ytwo);
    \end{pgfonlayer}
\end{tikzpicture}
	\caption{Testing environment: target for knowledge.} \label{fig:testing_env}
\end{subfigure}
\caption{Corridor environment for testing knowledge transfer.} \label{fig:transfer}
\end{figure}
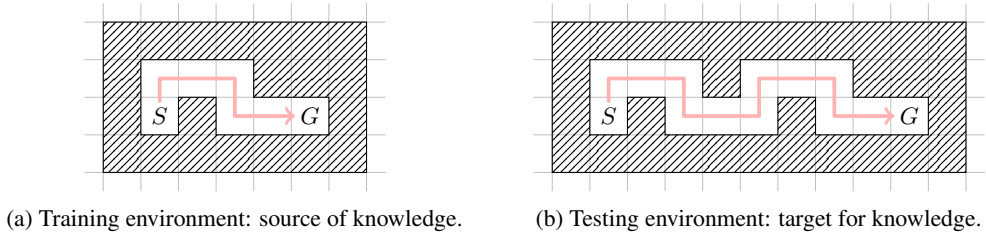

\Cref{fig:transfer} shows the corridor environment for testing knowledge transfer in RL agents. \Cref{fig:training_env} illustrates the training environment and \cref{fig:testing_env} the environment where the acquired knowledge is tested. In each time step, the agent receives a reward of $-1 $. If it reaches the goal position $G $, the current episode ends, its position is reset to the starting position $S $, and it receives a reward of $+10 $.

\emph{Objective interaction data} consists of the agent's absolute position as sensor input. Objective motor output determines whether it enters the cell to its north, east, south, or west in the next time step in case it is not occupied by a wall.\footnote{The agent does not receive additional punishment for running into walls.} The new part of the corridor in the larger testing environment contains qualitatively new objective data for the agent (i.e., coordinates with an unknown horizontal component). Therefore, it cannot benefit from the knowledge acquired in the training environment. It learns a new policy by successively overwriting the old one, as if it were in a completely new environment. 

\emph{Subjective interaction data} consists of the four surrounding cells as sensor input. Subjective motor output determines whether the agent turns 90 degrees to its right, to its left, or if it enters the cell in front of it (in case it is not occupied by a wall). Although the new part of the testing corridor is objectively different to the smaller training corridor, they subjectively appear identical to the agent. Subjective interaction data enables the agent to efficiently reapply its knowledge without even knowing so.

When knowledge transfer is the issue, the drawback of objective data becomes obvious. The drawback of subjective data is that any sufficiently complex environment presents behavior optimizing algorithms with POMDPs. To avoid semantically loaded representations, however, the set of world states cannot be provided a priori. POMDPs \emph{without} knowing the complete set of world states cannot be solved by any single policy. As our explicit goal is the reuse of representations in objectively different situations, however, it still seems reasonable to use subjective interaction data and cope with these problems as they occur.

Semantic load and knowledge transfer are essentially two stages of coping with the same problem. The problem of semantic load shows \emph{that} transferring information (i.e., semantics) might increase performance but it impairs autonomy at the same rate. Knowledge transfer, on the other hand, enables to investigate \emph{which} information can be transferred and \emph{how} to transfer it. The difference is, that the former is usually understood as pathological because the transferred knowledge originates \emph{in the designer,} while the latter is understood as beneficial because knowledge is transferred \emph{within the same system.}

In \cref{sec:intro} we have differentiated two perspectives on agent interaction that are of utmost importance for knowledge transfer. \Cref{sec:related} presents related work from the complementary fields of deep representation learning and hierarchical RL. In \cref{sec:overview}, we present the general components of our architecture and the type of information they exchange. \Cref{sec:implementation} sketches an instantiation of the architecture and the concrete machine learning methods applied. In \cref{sec:results} we present and discuss the results from testing our architecture in typical hierarchical RL gridworld environments. \Cref{sec:future} outlines subsequent research.

\section{Related Work} \label{sec:related}
Usually, the contrast between objective and subjective interaction is reduced to the difference between MDPs and POMDPs. To develop a machine learning algorithm from an appropriate model of the problem\-/to\-/be\-/solved, however, we see it fit not to take an observer's perspective but the agent's perspective on the problem. From the agent's perspective, there are no world states to begin with.

First and foremost, beliefs are \emph{identical} to the world perceived and influenced by the agent. To refrain from one's believes is an accomplishment of highly developed animals and by no means something to take for granted. POMDPs already imply a differentiation between belief and world states in separately modeling a probability distribution (i.e., a belief) and its targets (i.e., world states).

Therefore, POMDPs might be the perfect model for the problem of RL from an observer's perspective (i.e., for developing algorithms for objective interaction),\footnote{Although \cref{ssec:semantic} casts doubt on this.} to model learning \emph{as it is experienced,} however, requires to refrain from maintaining representations of objective world states. Instead, those states need to be constructed by the agent itself in a process of \emph{deep representation learning.}

\subsection{Deep Representation Learning}
An unbiased learner cannot rely on representations that have been provided by its designer. This insight dates back at least to the philosophical criticism that has been brought forward towards artificial intelligence in the early 80ies in the form of the \emph{Chinese room argument} \citep{Searle1980}, the \emph{epistemological frame problem} \citep{Dennett1981}, and, more explicitly, the \emph{symbol grounding problem} \citep{Harnad1990}. The semantic implications of representations have been recognized and are investigated in the field of \emph{representation learning.}

The most famous application of representation learning at the time being is in \emph{deep learning.} Deep learning is commonly performed with connectionist methods (e.g. restricted Boltzmann machines, auto\-/encoders, or artificial neural networks). Its eponymous feature, however, is not its class of methods but how they are employed.

One intuition behind deep learning is that data complexity can be reduced considerably by generating appropriate representations from the dynamics of interaction between a particular system and its environment.\footnote{Interestingly, this insight corresponds to recent developments in cognitive sciences that have been labeled as \emph{\enquote{embodied cognition}} \citep{Shapiro2011}.} Representation learning methods can be \enquote{stacked} by providing the representations generated in one layer to a higher\-/level instance of a similar representation learner. Therefore, in each layer, data complexity can be reduced more and more by recognizing and exploiting patterns. Data complexity is transformed into structural complexity of representations. As recognized patterns can be applied repeatedly, however, structural complexity only grows logarithmically with input complexity.

Connectionist variations of deep representation learners effectively treat the output of a layer as input for the layer above. The idea, however, can be generalized beyond connectionist implementations. Stacking any method that reduces data complexity by generating representations with less degrees of freedom (e.g. dimensionality reduction, principal component analysis, lossy, or loss\-/less compression) must eventually lead to a hierarchy of representations, with the most abstract at the top and the most concrete at the bottom.

If representations are generated not only from proto\-/features, sub\-/symbolic information, raw input, or however one may call the first data to enter the system, but also from \emph{previously learned representations,} then the system effectively generates layers containing abstract representations. A hierarchical organization of cognition comes naturally if a learning system is able to \emph{reuse} the representations in different layers. Benefiting from the same set of representations in objectively different situations is a case of knowledge transfer.

Hierarchical representations of a \emph{passive data space,} e.g. in image analysis \citep{Schmidhuber2014} or loss\-/less sequence compression \citep{Nevill-Manning1997}, have been proposed in various fields of computer science and with various applications. Approaches to learn the structure of hierarchical representations of an \emph{interactive data space} (i.e., a policy) go back at least to the early 90ies (\citealp{Schmidhuber1990}; \citealp{Schmidhuber1991}; \citealp{Feldkamp1998}), but only recently, \emph{RL} methods have been presented that are efficiently able to generate deep policies from scratch (\citealp{Mnih2013}; \citealp{Guo2014}). 

\subsection{Reinforcement Learning}
RL is concerned with optimizing behavior in unknown environments. Optimality is quantified as reward function that provides an evaluation feedback for the agent, indicating whether an action should be considered as good or bad. To achieve optimal behavior over time, the agent optimizes not the immediate but the \emph{cumulative} reward it expects to receive by following the current policy. This cumulative reward is discounted over future states because immediate rewards might matter more than those far in the future.

In general, such an evaluation can be estimated by remembering which action in which sensor state was followed by which discounted, cumulative reward. The future discounted, cumulative reward $R $ of unknown perception\-/action tuple $(s, a) \in S \times A $ can be predicted by estimating a value function $V $.
\begin{align} \label{eq:value}
V: S \times A \rightarrow R
\end{align}
Choosing a perception\-/action pair that maximizes the value function enables optimal behavior over a certain period of time.

The same action in the same state may not always lead to the same successor state. Non\-/stationary probability distributions can force the value function to adapt. More importantly, however, the same state\-/action pair might produce different states \emph{systematically.} To avoid conflating the rewards of those states, the value function can be augmented by past experience about state transitions in the observed environment. \emph{Model\-/based RL} explicitly models these state transitions.

Optimal models are identical to the transition function of the environment $\mathcal{T} $. Interpolating transition function $T $ in \cref{eq:trans} enables predicting the outcome of perception\-/action pairs that have not been observed before.
\begin{align} \label{eq:trans}
T: S \times A \rightarrow S
\end{align}

In model\-/based RL, instead of associating cumulative reward with perception\-/action pairs, immediate rewards are associated with sensor states. Knowledge about state transitions in the specific environment enables to choose attractive successor states via their estimated immediate reward \emph{and} the probability of reaching them with a particular action. Choosing an action that maximizes the estimated reward of the estimated successor states enables planned behavior.

More importantly, if a reliable model of the environment is available, the value of each perception\-/action tuple need not be tediously approximated by actually experiencing the future cumulative reward over all of its possible consequences. Instead, state transitions can be \enquote{simulated} arbitrarily far into the (estimated) future by exploiting the internal model.

Data structures which implement such a transition model are \emph{representations} of parts of the environment. Value and model function \cref{eq:value,eq:trans} determine the current interaction policy $\pi = (V, T) $ for the agent.

\emph{Hierarchical RL} explores hierarchical models of the environment. Most of recent research uses objective interaction data \citep{Sammut2011}. First and foremost reason is that objective data produces MDPs. If the agent is in a determinate sensor state, and any given action has (stochastically) predictable outcome (i.e., successor state and reward), transition and value functions can be approximated precisely.

Intuitively, transitions of abstract states in a high layer of the hierarchy occur less often than transitions in a low layer. Therefore, \emph{semi MDPs} are frequently used to model such time\-/sensitivity (\citealp{Thrun1995}; \citealp{McGovern1997}; \citealp{Dietterich1999}; \citealp{McGovern2001}; \citealp{Hengst2002}). This generalization of MDPs extends the model of state transitions by \emph{duration} as proposed by \citet{Sutton1999}.

The structure of hierarchical models, however, is mostly \emph{trained}:
 stochastic information is integrated into a fixed hierarchical structure. It is desirable to enable a RL agent to generate hierarchical representations \emph{autonomously,} on the one hand, to avoid the problem of semantically loaded representations and, on the other hand, to resolve the various ambiguities that result from subjective interaction in POMDPs. Deep representation learning methods can help to achieve this goal.




\section{Overview} \label{sec:overview}
In the following, we will present a rough outline of our approach to the grounding of deep models for RL. We present the general form of the functions we used. We implemented several agents according to this architecture. In this section, however, we present only the general structure such that the individual methods remain exchangeable.

We have shown that subjective interaction requires the disambiguation of seemingly identical states. To differentiate identical observations, additional information is necessary. This information is not directly accessible, it needs to be \enquote{uncovered in,} or \enquote{generated from,} observable data. This hidden information can be modeled as \emph{latent states.} In belief MDPs, the current world state is such a latent state. To avoid semantic load, however, we have determined that representations (i.e., the set of latent states) cannot be predefined but must be inferred from the dynamics of interaction between agent and environment alone. These dynamics can be stored in \emph{trajectories} or \emph{interaction histories.}\footnote{Although we regard both as equivalent (in contrast, see \citealp{Singh2000}), here, we will stick with the latter term, as it has already been established in connectionist RL \citep{Lin1993}. The former is more common in applications of \emph{dynamic systems theory} (e.g. \citealp{Guenter2007}).}

A history $h $ is a sequence of single \emph{experiences} $e $. In the literature, experiences usually consist of the 4-tuple $(s_t, a_t, s_{t+1}, r_{t+1}) $ with perception and action at time step $t $, and perception and reward at time step $t+1 $ \citep{Lin1993}. The perceived history $h $ can be recognized as an instance of a latent\-/state\-/specific model $M_l $, where $\forall l \in L.\; l \leftrightarrow M_l $ during latent state $l $.

Inspired by Peircean semiotics, our definition of representation consists of shape (i.e., a latent state $l $), content (i.e., a policy $\pi_l = (V_l, M_l) $ of value and model function), and reference (i.e., history $h_l $). A history is considered as the reference of a particular latent state, if the likelihood of the history being generated from the latent state's model exceeds a particular threshold.\footnote{Note that only \emph{linear} histories can be observed. Models, on the other hand contain \emph{nonlinear} information about \emph{possible} transitions, whereas histories only contain \emph{actual} transitions.} Latent states are the shapes of \emph{abstract representations} for situations in which their transitions hold. These situations are interaction histories.

\subsection{Learning Query Processes}
Transitions between abstract representations can be described by the \emph{query function.} The query function is analogue to the transition function $\mathcal{T} $ in MDPs. Notice, however, that MDPs assume a perception\-/action tuple to determine a single successor perception, whereas in an abstract representation, various queries from abstract representations might succeed and\slash or fail.
\begin{align} \label{eq:latent_query}
\mathcal{Q}: L \times L \rightarrow S \qquad S = \{\top, \bot \}
\end{align}

Choosing an abstract representation, and therefore its corresponding model, \emph{is not a MDP.} The reason is that inducing a policy is not as straight\-/forward as performing an action. Whether the model function of a queried latent state matches the subsequent history is not fully under control of the agent. Among abstract representations, there is no unambiguous distribution of responsibilities between agent and environment as it is among perceptions and actions.

In general, transitions of representations follow \emph{query processes} (QPs). A QP is defined by the 3-tuple $(L, \mathcal{Q}, \mathcal{R}) $. It contains the open set of latent states, the query function in \cref{eq:latent_query}, and the \emph{reward function} $\mathcal{R}: L \rightarrow R $, where $R = \mathbb{Z} $. The reward function in QPs is analogue to the reward function in MDPs.

An agent must be able to estimate whether a latent state can be induced from another latent state at all. To perform successful queries, therefore, the query function must be approximated by a \emph{model function.} This model function is analogue to the model function in \cref{eq:trans}. Optimal model functions are identical to the query function of the environment $\mathcal{Q} $.
\begin{align} \label{eq:model}
M: L \times L \rightarrow S \qquad S = \{\top, \bot \}
\end{align}

Abstract representations also need to be chosen according to the expected change in discounted, cumulative reward caused by their policy. This can be estimated by the \emph{abstract value function.} The abstract value function is analogue to the value function in \cref{eq:value}.
\begin{align} \label{eq:context_value}
V_l: L \rightarrow R
\end{align}

\Cref{eq:model,eq:context_value} determine the policy $\pi = (V_l, M) $ for solving QPs, just like \cref{eq:value,eq:trans} determine the policy $\pi = (V, T) $ for solving MDPs. The experiences in history $h $ enable to iteratively approximate value and transition function.\footnote{At this stage, the overall architecture resembles a \emph{hidden Markov model.} As more and more layers are added, however, the binary differentiation between latent and observable layers makes way for an unbounded hierarchy of stacked QPs. Also, notice that transitions between latent states are not determined by incoming information alone but they are also a means to inject information into the environment via actions.}

\subsection{Abstracting Query Processes}
Associating histories with models enables abstract representations of concrete observations in the form of latent states. These latent states model the agent's abstract perception of the environment: they represent the environment to the agent. A function to differentiate a perceived history according to the latent state  whose model matches best has the general form
\begin{align} \label{eq:abstract}
A: H \rightarrow L
\end{align}
where $H = [e_{t-n}, \ldots, e_t] $, $t $ is the current time\-/step, and $n $ is the length of the history. This function realizes an abstract representation of a history referenced by a the shape of a latent state via the model it contains. Therefore, we call it \emph{\enquote{abstraction function.}}

The environment responds to queries by either accepting or rejecting the queried state, depending on whether the interaction history that has been observed after querying a representation actually matches the according model function. If it does, the query is successful. Such a measure of \enquote{matching} must be integrated in the abstraction function, such that only latent states are returned that are associated with models compatible to the observed history. 

Each abstract representation provides an interaction policy that consists of value and model function. To effectively reuse policies, they must be efficiently retrievable. The general form of a retrieval function must be inverse to $A $: not from observed histories to representations, but from representations to concrete interaction policies. This \emph{application function} has the following form:
\begin{align}
A': L \rightarrow \Pi
\end{align}

Environment and agent determine \emph{together} where the transition from the current representation leads to. Consequentially, the intension to induce a particular successor representation $l_q $ may fail, dependent on whether the current history which has been evoked by the queried policy $h_t^{\pi_q} $ is actually compatible to this policy. Whether a query succeeds or fails is determined in \cref{eq:query_success}.
\begin{align}
h_t^{\pi_q} & = [e_{t-n}, \dots, e_t] \nonumber \\
l_t & = A(h_t^{\pi_q}) \nonumber \\
\mathcal{Q}(l_{t-n-1}, l_q) & \leftrightarrow l_t = l_q \label{eq:query_success}
\end{align}

\subsection{Stacking Query Processes}
There are obvious parallels between observable and latent states. Value functions are applicable straight\-/forwardly in both. In both cases, choices influence the future reward. The forms of the model for observable state transitions in \cref{eq:trans} and the model for representation transitions in \cref{eq:model}, however, differ. MDPs regard \emph{observable states as conditional on a perception\-/action tuple,} whereas QPs regard \emph{the success of a query as conditional on a tuple of two consecutive latent states.} As we have seen, latent states cannot be differentiated into perception and action. The general form of the model function of MDPs in \cref{eq:trans} can therefore not be applied to them. 

To enable an unbounded hierarchy, it is desirable to unify the model functions in each layer. QPs can be stacked. Abstract model functions can be generated which describe the transitions of latent states. Latent states, on the other hand, contain low\-/level model functions themselves. To be grounded in observable states, however, concrete sensorimotor interaction must also be described by QPs.

If we perform some modifications to what we understand as an observable state, the model function of QPs in \cref{eq:model} can be applied to observable states. We have shown that observable states in RL are commonly associated with the sensor states of an agent. Its actions, on the other hand, are also observable, but essentially different. Action $a_t $ and perception $s_t $ are separate. To translate MDPs into QPs, action and perception need to be integrated. We introduce \emph{sensorimotor states} by simple concatenation of the last action and the current observation: $x = (a_{t-1}, s_t) $ and $x \in SM $ where $SM = A \times S $.

Sensorimotor states can be queried like latent states. A sensorimotor state, however, does not imply an interaction policy, because sensorimotor states are the lowest layer of interaction. Instead, querying a sensorimotor state implies the execution of the part of the state the agent has control over: the motor component. The agent has only partial influence on the success of a query. The sensor component is controlled by the environment. If the sensorimotor state in the query contains another sensor component than the agent's perception in the next time step, the query will fail.

Modifying observable states also implies modifying experiences. For QPs we differentiate two types of experiences. One contributing to the value function and one contributing to the model function. \emph{Value\-/related experiences} consist of a sensorimotor state and the simultaneously received reward $e_v = (x_t, r_t) $. \emph{Model\-/related experiences} consist of an observed sensorimotor state, a queried sensorimotor state, and whether the query was successful $e_m = (x, q, s) $ where $s \in \{\top, \bot\} $ and $x, q \in SM $.





\section{Implementation} \label{sec:implementation}
In the following, we describe how we implemented the above functions and related methods from conventional RL. The \emph{value function} estimates the discounted, cumulative reward. The \emph{model function} estimates which sensorimotor states can be queried. The query function $\mathcal{Q} $ in QPs replaces the transition function $\mathcal{T} $ in MDPs. Finally, the \emph{abstraction function} generalizes histories of experiences of sensorimotor states into high\-/level representations. The \emph{application function} merely remembers which value and model functions are associated with which abstract state. Therefore, we will not cover it in more detail. We will present practical methods to approximate these functions.\footnote{In the present research, we evaluate QPs as a means for \emph{grounding} a hierarchical model for RL. Our ongoing research investigates the possibility to effectively generate such a model.}

\subsection{Value Function}
RL literature provides numerous methods for approximating a value function. Two of the most prominent ones are \emph{Q-learning} (originally \citealp{Watkins1989}) and \emph{SARSA-learning} (interestingly, originally in a connectionnist context \citealp{Rummery1994}). The former generates a value function which describes the \emph{optimal interaction policy} (i.e., offline learning), the latter generates a value function which describes the \emph{current interaction policy} (i.e., online learning).

Consider an agent that performs its (arbitrarily chosen) actions only with a certain probability. In some cases it performs a random action instead. The task is to learn to walk the shortest possible path along a straight cliff. Q-agents learn to walk close to the edge, because it is the shortest path with the most cumulative reward. If a random action occurs, however, they might drop, suffering large amounts of negative reward. Q-learning performs offline learning.

Online policy learners, on the other hand, learn from actually performed actions. In the cliff walking example, they learn to follow an arc that keeps a safe distance to the edge. Any autonomously learning agent needs to perform random actions to explore unknown environments. For life\-/long learning, therefore, online learning seems preferable.

For this reason we adopted SARSA-learning. The original update function for separate sensor and motor states in SARSA-learning is as follows.
\begin{align} \label{eq:sarsa}
V(s_{t-1}, a_{t-1}) \leftarrow V(s_{t-1}, a_{t-1}) + \alpha [r_t + \gamma V(s_t, a_t) - V(s_{t-1}, a_{t-1})] \qquad \alpha, \gamma \in [0, 1]
\end{align}

This iterative process is parametrized with learning factor $\alpha $ and discount factor $\gamma $. The learning factor determines the readiness to overwrite previous experiences with new ones. The bigger $\alpha $ is, the more weight new value\-/related experiences have in comparison to older ones. The discount factor $\gamma $ determines the importance of future rewards.

Q- and SARSA-learning use MDPs as a model. Therefore, both separate sensor and motor states. They are, however, easily adaptable to accommodate for an element of the set of all sensorimotor states $x \in SM $. We modified the SARSA\-/value update in the following way. Notice that in our modification, the simultaneously received reward is used to update the value of a state, instead of the next reward as in \cref{eq:sarsa}.
\begin{align}
V_l(x_{t-1}) \leftarrow V_l(x_{t-1}) + \alpha [r_{t-1} + \gamma V_l(x_t) - V_l(x_{t-1})] \qquad \alpha, \gamma \in [0, 1]
\end{align}

If the task is segmented into episodes, the expected cumulative reward has an upper bound (i.e., the cumulative reward to be expected during the remainder of the current episode). If the task is ongoing, however, the value function must be bounded by normalization.
\begin{align} \label{eq:myvalueupdate}
V_l(x_{t-1}) \leftarrow \frac{V_l(x_{t-1}) + \alpha [r_{t-1} + \gamma V_l(x_t) - V_l(x_{t-1})]}{1 + \alpha} \qquad \alpha, \gamma \in [0, 1]
\end{align}

Q- and SARSA-learning are used in model\-/free RL. The interaction policy followed by a Q- or SARSA-agent is determined by the value function: the agent performs the one action that maximizes the value function. In model\-/based RL, however, the reward estimate is only one component responsible for a particular policy.

\subsection{Model Function}
The environment can be modeled by approximating the transition function for observable states. Transition probabilities are increased for actually observed transitions, see \cref{eq:obs_trans}, and decreased for all other transitions, see \cref{eq:oth_trans}.
\begin{align}
T(s_{t-1}, a_{t-1}, s_t) \leftarrow & \; T(s_{t-1}, a_{t-1}, s_t) + \alpha (1 - T(s_{t-1}, a_{t-1}, s_t)) \label{eq:obs_trans} \\
\forall s' \in S \wedge s' \neq s_t. T(s_{t-1}, a_{t-1}, s') \leftarrow & \; T(s_{t-1}, a_{t-1}, s') + \alpha (0 - T(s_{t-1}, a_{t-1}, s')) \label{eq:oth_trans}
\end{align}

Estimates of state transitions enable to plan behavior. Traditionally, transition functions realize models that enable to estimate transition probabilities from one sensor state via action to another sensor state. They are usually not combined with value functions that provide estimates of \emph{cumulative} reward, because the expected cumulative reward can be \enquote{simulated} with an appropriate model. \Cref{eq:reward} shows how model\-/based approaches rather estimate the \emph{immediate} reward of perception\-/action pairs.
\begin{align}
R(s_t, a_t) \leftarrow & \; R(s_t, a_t) + \alpha (r_t - R(s_t, a_t)) \label{eq:reward}
\end{align}

Given a comprehensive model of state transitions and expected rewards, model\-/based reinforcement agents are able to chose the action that maximizes the sum of expected rewards weighted by the probability of expected future states. The probability of future states is estimated by the model function. By recursively retrieving the estimated value $V $ of potential successor states, model\-/based agents are able to determine values that take states into account which are arbitrarily far in the future.
\begin{align}
V(s, a) = R(s, a) + \gamma \sum_{s'}^S T(s, a, s') \max_{a'}^A V(s', a') \label{eq:model_value}
\end{align}

The recursive call in \cref{eq:model_value} enables to use an existing model for estimating future cumulative reward, instead of visiting each state several times to cover all of its possible consequences like in in \cref{eq:sarsa} of model\-/free RL.

The downside is that a recursion necessitates the definition of a termination condition. Introducing conditions always comes at the risk of corrupting universality. Secondly, the value function has to be evaluated several times in each iteration and, lastly but most importantly, an approximately correct model of the environment needs to be already available to estimate the value of states.

By incorporating the iterative value update in \cref{eq:myvalueupdate}, we are able to avoid a potentially expensive recursion. Our architecture tries to exploit both: a model of the environment \emph{and} an estimate of the long\-/term attractiveness of a state. With each query it considers only those sensorimotor states that the model deems possible. Among those states, the value function enables to choose the most promising ones.

Our approximation of the model function can be seen in the following. The value of $s $ indicates whether the environment accepted the last query $q_{t-1} \in SM $ or not (i.e., whether $q_{t-1} $ is identical to the current state $x_t $). It follows that the \enquote{inducibility} of query $q $ from sensorimotor state $x $ can be determined by \cref{eq:induce}.
\begin{align}
I_l(x_{t-1}, q_{t-1}) \leftarrow I_l(x_{t-1}, q_{t-1}) + \alpha (s - I_l(x_{t-1}, q_{t-1})) \qquad s =
\begin{cases}
	1,& \text{if } q_{t-1} = x_t \\
	0,& \text{otherwise}
\end{cases} \label{eq:induce}
\end{align}

\subsection{Action Selection}
If the value function enables to evaluate perception\-/action tuples, then action selection simply returns the action $a $ from the a tuple that maximizes the estimated value among all perception\-/action tuples. Drawbacks are costs linear to the sum of the number of perceptions and the number of actions. \Cref{eq:dec_action} shows the \emph{decision function} in conventional model\-/based and model\-/free RL during perception $s_t $.
\begin{align}
D(s_t) = a \qquad V(s_t, a) = \max_{s_t, a'}^{S \times A} V(s_t, a') \label{eq:dec_action}
\end{align}

Instead of the probability of successor states, query selection must consider the \enquote{inducibility} of potential successor states. The approach also performs a simple maximization. Its specific form, however, is quite different from conventional action selection in RL. The decision function realizes query selection determined by the value function and the approximation of the model function.
\begin{align}
D(x_t) = x \qquad V_l(x) = \max_{x'}^{X_i} V_l(x'), \quad \forall x_i \subset X_i.\; I(x_t, x_i) \geq c \quad c \in [0, 1] \label{eq:query}
\end{align}

\Cref{eq:query} enables to select sensorimotor \emph{and} latent states alike. The state $x $ can either be a sensorimotor or a latent state. For each query, first, the set of most inducible states with a certainty above $c $ is selected. From this set, a query is selected that maximizes the value function. Both parts of the current policy $\pi $, value function $V $ and model function $M $, are determined by the currently active latent state $l $. This latent state, on the other hand, is subject to simple QPs, just as low\-/level sensorimotor states are. From the explications above the architecture in \cref{fig:architecture} emerges.

\begin{figure}
\centering
\begin{minipage}{.5\linewidth}
	\centering
\pgfdeclarelayer{background}
\pgfdeclarelayer{foreground}
\pgfsetlayers{background,main,foreground}


\tikzstyle{state}=[draw, fill=green!20,
    text centered,circle,drop shadow]

\tikzstyle{sensor}=[draw, fill=blue!20,
    text centered,drop shadow]

\tikzstyle{wa} = [sensor,fill=red!20,
    rounded corners, drop shadow]

\begin{tikzpicture}[scale=.7]
    \node (wa) [wa]  {abstract layer};
    \path (wa.south)+(-2.5,-6) node (asr1) [state] {$sm_t $};
    \path (wa.south)+(-2.5,-3.5) node (asr2) [sensor] {$h_t = [sm_{t-n}, \dots, sm_t] $};

    \path (wa.south)+(-2.5,-1.5) node (dots) [state] {$l_t $};
    \path (wa.south)+(+2,-1.5) node (asr3) [state] {$l_q $};    

    \path (wa.south)+(+2,-3.5) node (policy) [sensor] {$\pi = (V_l, M) $};    
    \path (wa.south)+(+2,-6) node (query) [state] {$sm_q $};    
   
    \path [draw, ->] (asr1.north) -- node [right] {$D^1 $} 
        (asr2.south) ;
    \path [draw, ->] (asr2.north) -- node [right] {$A $} 
        (dots.south);
    \path [draw, ->] (dots.north) -- node [right] {$D^2 $} 
        (wa.south -| wa.west);
    \path [draw, ->] (wa.south -| wa.east) -- node [right] {$D^2 $} 
        (asr3.north);

    \path [draw, ->] (asr3.south) -- node [right] {$A' $} 
        (policy.north);
    \path [draw, ->] (policy.south) -- node [right] {$D^1 $} 
        (query.north);
               
    \path (wa.south)+(0,2) node (asrs) {base layer};
  
    \begin{pgfonlayer}{background}          
        \path[fill=yellow!20,rounded corners, draw=black!50, dashed]
            (asr2.west |- wa.north)+(-1,+1) rectangle (4.5,-4.5);
            
    \end{pgfonlayer}

\end{tikzpicture}
	\captionof{figure}{Overview of base layer of the architecture and the recursive integration of further layers.}
	\label{fig:architecture}
\end{minipage}%
\begin{minipage}{.5\linewidth}
	\centering
	\begin{tikzpicture}[scale=.4, even odd rule]\footnotesize
    \pgfmathsetmacro{\xone}{-.5}
    \pgfmathsetmacro{\xtwo}{15.5}
    \pgfmathsetmacro{\yone}{-.5}
    \pgfmathsetmacro{\ytwo}{15.5}
    \pgfmathsetmacro{\xs}{5}
    \pgfmathsetmacro{\ys}{5}

    \begin{scope}<+->;
        \draw[step=1cm,gray,very thin, opacity=.5] (\xone,\yone) grid (\xtwo,\ytwo);
    \end{scope}

    \begin{scope}
        \draw[dashed,line width=.05cm,color=red!30,cap=round,join=round,->] (4.9,4.5)--(10.5,4.5)--(10.5,10.1);
        \draw[line width=.05cm,color=red!30,cap=round,join=round,->] (4.5,4.9)--(4.5,10.5)--(10.1,10.5);
        \draw[dashed,line width=.05cm,color=red!30,cap=round,join=round] (7.5,4.5)--(7.5,10.5);
        \draw[dashed,line width=.05cm,color=red!30,cap=round,join=round] (4.5,7.5)--(10.5,7.5);

        \draw[black, pattern=north east lines] (0, 0) rectangle (15, 15) (1,1) rectangle (14,14);

        \draw[black, pattern=north east lines] (2, 2) rectangle (4, 4);
        \draw[black, pattern=north east lines] (5, 2) rectangle (7, 4);
        \draw[black, pattern=north east lines] (8, 2) rectangle (10, 4);
        \draw[black, pattern=north east lines] (11, 2) rectangle (13, 4);

        \draw[black, pattern=north east lines] (2, 5) rectangle (4, 7);
        \draw[black, pattern=north east lines] (5, 5) rectangle (7, 7);
        \draw[black, pattern=north east lines] (8, 5) rectangle (10, 7);
        \draw[black, pattern=north east lines] (11, 5) rectangle (13, 7);

        \draw[black, pattern=north east lines] (2, 8) rectangle (4, 10);
        \draw[black, pattern=north east lines] (5, 8) rectangle (7, 10);
        \draw[black, pattern=north east lines] (8, 8) rectangle (10, 10);
        \draw[black, pattern=north east lines] (11, 8) rectangle (13, 10);

        \draw[black, pattern=north east lines] (2, 11) rectangle (4, 13);
        \draw[black, pattern=north east lines] (5, 11) rectangle (7, 13);
        \draw[black, pattern=north east lines] (8, 11) rectangle (10, 13);
        \draw[black, pattern=north east lines] (11, 11) rectangle (13, 13);

        \draw[black] (4.5, 4.5) node {$S$};
        \draw[black] (10.5, 10.5) node {$G$};
    \end{scope}

    \begin{pgfonlayer}{background}
	    \fill [white] (\xone, \yone) rectangle (\xtwo, \ytwo);
    \end{pgfonlayer}
\end{tikzpicture}
	\captionof{figure}{Labyrinth environment.}
	\label{fig:motmem}
\end{minipage}
\end{figure}

\section{Results} \label{sec:results}
The more abstract representations get, the more blurry the line between action and perception becomes. In contrast to MDPs, QPs can be applied to abstract state transitions. By continuously performing queries for desirable states, an agent can learn transition probabilities between these states. Once the query function $\mathcal{Q} $ has been sufficiently approximated, the agent can choose those successor states that promise to maximize a normalized variant of expected future cumulative reward. This behavior effectively realizes a value optimizing policy in a particular QP.

Although both, Markovian policy and query policy, converge in the small corridor environment from \cref{fig:training_env} after 14 episodes, QPs take much longer to complete an episode.\footnote{In all experiments we set $\alpha = .5 $ and $\gamma = .5 $ with an initial optimistic value of $5 $.} The performance of the query policy suffers from condensing sensor and motor states. The decline in performance can be explained by the increased number of sensorimotor states. Whereas a Markovian model function needs to keep track of $|S \times A \times S| $ entries ($432 $ values in the small corridor environment with subjective interaction), the query model function needs to track $2 * |S \times A |^2 $ entries ($2592 $ values, respectively). The value function $V $ in model\-/based Markovian policies contains $|S| $ values ($12 $ in our example), the value function in query policies contains $|S \times A |$ entries ($36 $, respectively).


Consider, however, that subjective policies do not grow with new objective input. As long as objectively unknown input can be interpreted as subjectively known, the policy size remains unchanged. Of course, this advantage of subjective interaction also holds for Markovian policies. As we have shown, however, the distinction between perception and action effectively prevents any straight\-/forward application of Markov policies to latent states. Any ambitions towards a hierarchy of representations with MDPs will first have to solve the problem of how to integrate perception and action.

\begin{figure}
	\centering
	\begin{subfigure}[b]{.48\textwidth}
		\centering
		\pgfplotsset{max space between ticks=50pt}
\tikzset{every mark/.append style={scale=.5}}

\begin{tikzpicture}
\begin{axis}[
		scaled ticks=false,
		height=5cm,
		width=\textwidth,
		ymax=0,
		ymin=-3000,
        xmax=60,
        xmin=0,
		legend style={at={(.2, .7)}, anchor=west},
        ]

    \addlegendentry{objective Markov}
    \addplot+[brown]
        table[x=episode,y=reward,y error=error]
        {data/abs_compare.csv};

    \addlegendentry{subjective Markov}
    \addplot+[blue]
        table[x=episode,y=reward,y error=error]
        {data/markov_fail.csv};

    \addplot+[red, no markers] coordinates {(0,-1) (60,-1)}; 
    
\end{axis}
\end{tikzpicture}
		\caption{Performance of Markov policies.} \label{fig:markov_fail}
	\end{subfigure}
	~
	\begin{subfigure}[b]{.48\textwidth}
		\centering
		\pgfplotsset{max space between ticks=50pt}
\tikzset{every mark/.append style={scale=.5}}

\begin{tikzpicture}
\begin{axis}[
		scaled ticks=false,
		height=5cm,
		width=\textwidth,
		ymax=0,
        xmax=30,
        xmin=0,
		legend style={at={(.3, .2)}, anchor=west},
        ]

    \addlegendentry{subjective query}
    \addplot+[black]
        table[x=episode,y=reward,y error=error]
        {data/query_success.csv};

    \addlegendentry{optimal policy} 
    \addplot+[red, no markers] coordinates {(0,-1) (30,-1)};

    \addplot+[brown]
        table[x=episode,y=reward,y error=error]
        {data/abs_compare.csv};

\end{axis}
\end{tikzpicture}
		\caption{Performance subjective query policy.} \label{fig:query_success}
	\end{subfigure}
	\caption{Cumulative reward per episode in the labyrinth environment.}
\end{figure}
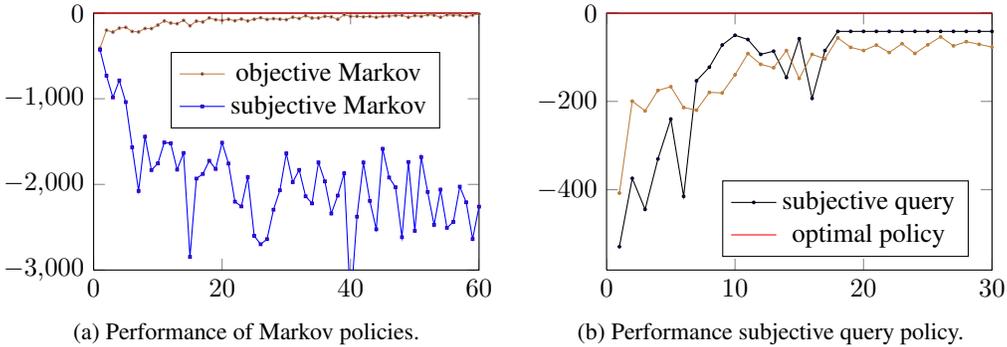

\Cref{fig:motmem} illustrates the labyrinth environment. Reward and interaction conditions are identical to the corridor setting. In contrast to the small corridor environment, however, the labyrinth cannot be solved by a Markovian policy given subjective interaction data. To reach the goal, an agent has to go straight at some crossroads but turn at others. Without further means of differentiation, subjective Markov policies can only perform one of both each time. \Cref{fig:markov_fail} shows that the labyrinth cannot be solved by MDPs with subjective interaction data.

The policy of the objective Markov agent converges at the optimum after roughly 140 episodes. \Cref{fig:query_success} shows that the  policy of the subjective query agent converges at sub\-/optimal performance after only 18 episodes. A query policy is able to differentiate the identical perceptions at crossroads according to the last performed action: it develops a strategy to solve the labyrinth.

The following policy is one of the successful, sub\-/optimal policies developed by a query agent. The policy has been condensed, such that the underlying search strategy becomes obvious. The conditionals to the left are sensorimotor states, consisting of the last action and the current perception.
\begin{itemize}
\item (turn, a wall to the right): turn left
\item (turn, at a crossroad): move forward until one of the following holds.
\item (forward, at a crossroad): turn left
\item (forward, facing a wall): turn left
\end{itemize}

By storing the last motor activation, query policies with sensorimotor states realize a sort of short\-/term memory. Subjective single\-/layer query policies can solve tasks that cannot be solved by subjective single\-/layer Markovian policies. Such a \enquote{graceful decent} of performance is highly desirable if the relevant states of the task are not known in advance. Even if the agent is not able to develop a policy that is optimal \emph{from an observer's point of view,} it can still produce solutions that take its own limitations into account.

\section{Future Work} \label{sec:future}
The greatest challenge for extruding a grounded query policy into a hierarchical architecture is to find an appropriate abstraction function $A $ that reliably and repeatedly provides the same models given only linear histories of them. The fact that the dimensionality of the history is always smaller or equal to the dimensionality of the model complicates reliable comparisons. The history might not contain the information relevant for deciding whether is corresponds to one model or another.

Additional information needs to be acquired to resolve ambiguities. It is clear, however, that this information cannot be extracted from the structure of history or model. Representation learning might be able to generalize histories, but any generalization that considers only the structure of history and model is confronted with underdetermined exemplars: it cannot be decided which one of a structurally equally similar set of models a history belongs to. In cognitive science, a similar situation presents a problem in explaining the universality of human object recognition. It is known as \emph{the problem of vanishing intersections} \citep{Harnad1990}.

A hierarchical architecture like in \cref{fig:architecture}, however, can estimate the model of an observed history from information beyond sheer structure. The transitions of policies, and therefore the transitions of models, that are motivated by observing a particular history are at the same time abstract state transitions in the layer above. To be able to model and predict abstract state transitions (e.g. as QPs) implies the ability to model and predict lower policy transitions.

The top layer effectively provides prior probabilities for policy transitions. Even without any useful structural information (i.e., no significant intersection of history and model), the likelihood for a particular history for having been generated by a particular model can be acquired by previously experienced abstract state transitions. This realizes an \emph{inductive expectation bias.}

QPs provide the formal grounds for augmenting base layer query selection $D^1 $ with an unbounded number of more abstract and less frequent query selections $D^m $ where $m = [1,\infty] $. Layers grow as the acquired knowledge is no longer applicable in every situation. The general case, however, is to assume that every environment can be described in terms of a QP. No sooner than when this assumption is violated, new data structures (i.e., representations) are generated. Until then, available knowledge is transfered onto every objectively new situation \emph{per se.}

\bibliography{iclr2015}
\bibliographystyle{iclr2015}

\end{document}